\documentclass{article}

\usepackage[numbers]{natbib}
 \usepackage[final]{nips_2017}

\usepackage[utf8]{inputenc} % allow utf-8 input
\usepackage[T1]{fontenc}    % use 8-bit T1 fonts
\usepackage{hyperref}       % hyperlinks
\usepackage{url}            % simple URL typesetting
\usepackage{booktabs}       % professional-quality tables
\usepackage{amsfonts}       % blackboard math symbols
\usepackage{nicefrac}       % compact symbols for 1/2, etc.
\usepackage{microtype}      % microtypography
\usepackage{graphicx}

\title{Uncertainty Estimates for Efficient Neural Network-based Dialogue Policy Optimisation}

\author{
  Christopher Tegho$^*$ \\
  Department of Engineering\\
  Cambridge University\\
  \texttt{christegho@gmail.com} \\
   \And
  Pawe\l{} Budzianowski$^*$ \\
  Department of Engineering\\
  Cambridge University\\
  \texttt{pfb30@cam.ac.uk} \\
   \And
 Milica Ga{\v s}i{\'c} \\
  Department of Engineering\\
  Cambridge University\\
  \texttt{ mg436@cam.ac.uk} \\
}

\begin{document}
\maketitle
{\let\thefootnote\relax\footnotetext{* Both authors contributed equally.}}

\begin{abstract}
In statistical dialogue management, the dialogue manager learns a policy that maps a belief state to an action for the system to perform. Efficient exploration is key to successful policy optimisation. Current deep reinforcement learning methods are very promising but rely on $\varepsilon$-greedy exploration, thus subjecting the user to a random choice of action during learning. 
Alternative approaches such as Gaussian Process SARSA (GPSARSA) estimate uncertainties and are sample efficient, leading to better user experience, but on the expense of a greater computational complexity.
This paper examines approaches to extract uncertainty estimates from deep Q-networks (DQN) in the context of dialogue management.
We perform an extensive benchmark of deep Bayesian methods to extract uncertainty estimates, namely Bayes-By-Backprop, dropout, its concrete variation, bootstrapped ensemble and $\alpha$-divergences,	
combining it with DQN algorithm.  

%, and show that dropout is a promising simple method worth further investigation.
%We find that BBQN achieves faster convergence to an optimal policy than any other method, and reaches performance comparable to the state of the art, but without the high computational complexity of GPSARSA.  %We also implement $\alpha$-divergences, variational dropout, and minimizing the negative log likelihood as other means to extract uncertainty estimates from DQN, and compare performance to BBQN and DQN. This work is carried within in the Cambridge University Engineering Department dialogue systems toolkit, CUED-pydial.
\end{abstract}

\section{Introduction}
\label{sec:intro}
%Spoken Dialogue Systems (SDSs) allow human users to interact with computers through speech. SDSs have became a common deployment in the speech interfaces in mobile phones, and are gaining greater commercial use. 
Statistical approaches to dialogue modelling allow automatic optimisation of the Spoken Dialogue Systems (SDS) \cite{young2002talking}.
A SDS is typically designed according to a structured ontology (or a database schema), which defines the domain that the system can talk about. The domain is presented using slots, which are variables the user can either specify or ask about in the domain. 

The system also comprises various statistical components. This includes a spoken language understanding module, which takes a sentence as input and gives a dialogue act as output \cite{henderson2015discriminative}, (e.g. requesting information, informing constraints, saying good-bye), and a slot-value pair that specify arguments of the act. For example \texttt{inform(hotel=expensive)} is a dialogue act of type \texttt{inform}, where the user is informing the system that they would like to constrain their search to luxury venues. The other components of a SDS include a dialogue belief state tracker that predicts user intent and track the dialogue history, a dialogue policy to determine the dialogue flow, and a natural language generator to convert conceptual representations into system responses.

The POMDP framework mitigates the problem of noisy estimates of spoken language understanding by maintaining a distribution over all possible hypothesises, which is called the belief state \cite{young2013pomdp}.
A dialogue policy is then employed to map the belief state into an appropriate system action at every dialogue turn. The ability to generalise across different noise levels is essential for successful dialogue policy operation.

%, this approach has several shortcomings:
%Dialogue corpora can be used to "mimic" the dialogue policy. However, this approach has several shortcomings:
%firstly
Although supervised learning from dialogue corpora can be used to learn human decision-making from data, the action selection does not take any future outcomes of the dialogue into considersation, which leads to a sub-optimal behaviour.
%the effect of selecting an action on the future course of the dialogue is not considered and this may result in sub-optimal behaviour; 
%moreover, the reward is often delayed in time, making one-turn optimisation inappropriate.
%secondly, the single-turn optimisation in supervised learning may not be appropriate because the feedback (or reward) is usually delayed in time.
As an alternative, Reinforcement Learning (RL), which maximizes the expected sum of rewards received over the course of a dialogue, is a much suitable method for learning dialogue policies \cite{levin2000stochastic}. The reward in this case measures the degree to which the dialogue is successful. 
%That is why, Reinforcement leaning (RL) is used to learn a policy over a dialogue that maximizes the expected sum of rewards received after visiting a state~\cite{levin2000stochastic}, and an action-state value function $Q$ can be computed for this purpose.

%For dialogue systems, the state-action space is large, and a Q-function approximation is necessary. 
%To explore the environment, an $\varepsilon$-greedy policy can be employed.

%Efficient exploration is as key here as we want to train the system by direct interactions with the real users.
Due to the need of learning with real users through online interactions, an efficient exploration of the state-action space is critical.
The Q-function of each state-action pair can be augmented with an estimate of its uncertainty to guide exploration to achieve higher performance and efficient learning \cite{daubigney2011uncertainty}.
Uncertainty estimates in the policy allow the system to generalise across different noise levels and mitigate errors incurred by speech recognition, therefore resulting in a more robust dialogue manager.
%Moreover, the ability to generalise across different noise levels is essential for successful dialogue policy management.
%Automatic speech recognisers often provide inaccurate translation in noisy environments.

Gaussian Processes (GPs) provide an explicit estimate of uncertainty, but are computational intensive and preclude the use for large action spaces \cite{gasic2014gaussian}.
% which helps to overcome the problems mentioned above, it is nevertheless a computational intensive algorithm and precludes the use for large action spaces.
%Gaussian Processes (GPs) provide an explicit estimate of uncertainty~\cite{gasic2014gaussian}, overcoming the above-mentioned problems. 
%However, GPSARSA is a computational intensive algorithm, 
%requires computing the inverse of a Gram matrix ${K}$ for determining the predictive posterior and estimating $Q$ at new locations, 
%precluding its use for large action spaces.
Deep neural network~(DNNs), on the other hand, scale much better with data and are computationally less expensive than GPs.
Many studies have shown that they are suitable for dialogue management tasks~\cite{cuayahuitl2015strategic,fatemi2016policy,williams2017hybrid,su2017sample}.
However, their application in noisy environments is relatively under-explored \cite{lipton2016efficient,chen2017agent}.
In this paper, we perform a benchmark of uncertainty estimates for a dialogue domain using Bayesian deep learning \cite{blundell2015weight,osband2016deep,li2017dropout,gal2017concrete}. We experiment with and without noise added to the simulated user input to examine generalization capabilities of different approaches and compare it to the state-of-the-art GPSARSA algorithm. The Bayes-by-Backprop algorithm achieves the best performance among the neural networks approaches.
\section{Uncertainty estimates in deep reinforcement learning}
\subsection{Deep Q-learning}
In the case of value-based reinforcement learning, we approximate the expected discounted sum of the future rewards received over the course of a dialogue given an action $a$ in a belief state $b$: 
$$
Q_t(b, a)= \mathbb{E}_{\pi} \{ r_{t} + \gamma r_{t+1} + ... ~~ | b_t=b, a_t=a\},
$$
where $r_t$ is the one-step reward received at a given time $t$. 

The Deep Q-Network (DQN) algorithm models action-value function $Q(b,a)$ using a deep neural network $\hat{Q}(b,a; \theta)$, with weight vector $\theta \in \mathbb{R}^{n}$, where we iteratively improve the prediction by minimizing the following loss:
$$\label{eq:dqn}
L({\theta}_t) = \mathbb{E}\left[(y_t - \hat{Q}(b_t,a_t;\theta_t))^2 \right],
$$
where the targets $y_t$ are:
$$ y_t =r_t + \gamma \max_{a'} \hat{Q}(s_{t+1}, a';\theta_t).$$
At testing time the system responds with an action that yields the highest Q-value for a given belief state $b$.
%The learning can be greatly improved by using the experience replay buffer that stores previous episodes \cite{lin1992self}.
%Value-based methods usually approximate $Q$-value function which has the form:
%$Q^{\pi}(b, a) = \mathbb{E}_{\pi} \{ R_{t} + \gamma R_{t+1} + ... ~~ | b_t=b, a_t=a\}.$
%We model action-value function using a deep neural network where we iteratively improve our guess by minimizing the loss:
%\begin{equation}\label{eq:dqn}
%\begin{aligned}
%L(\boldsymbol{\theta}_t) = \mathbb{E}\left[(y_t - \hat{Q}^{\pi}(b_t,a_t;\boldsymbol{\theta}_t))^2 \right]
%\end{aligned}
%\end{equation}
%where targets are 
%$ y_t =r_t + \gamma \max_{a'} \hat{Q}^{\pi}(b_{t+1}, a';\boldsymbol{\theta}_t)$ and 
%the expectation is usually taken with respect to $\epsilon$-greedy policy \cite{blundell2015weight}.

\subsection{Uncertainty estimates in neural networks}
To obtain uncertainty estimates from a neural network, Bayesian neural networks (BNNs) can be employed \cite{neal2012bayesian}. %BNNs are standard neural networks with prior probability distributions placed over the weights. 
Instead of having single fixed value weights $w$ in the neural networks, all weights are represented by probability distributions over possible values given observed dialogues $\mathcal{D}$, $P(w| \mathcal{D})$. Uncertainty in the hidden units allows the expression of uncertainty about predictions \cite{blundell2015weight}. 

For exploration, Thompson sampling is used instead of $\epsilon$-greedy, which consists of performing a single stochastic forward pass through the network every time an action needs to be taken. 
%The posterior distribution of the weights given 
%the training data, $P(w|\mathcal{D})$, is calculated with Bayesian inference. 
The $Q$-values given the input belief state $b$ are given by:
\begin{equation}
P(Q,b) = \mathop{\mathbb{E}}_{P(w|\mathcal{D})}[P(Q|b, w)].
\end{equation}
Taking an expectation under the posterior distribution is equivalent to using an ensemble of an uncountably infinite number of neural networks, which is intractable \cite{blundell2015weight}. We have to resort to sampling-based or stochastic variational inferences. 

\subsection{Benchmarked algorithms}
We used in this benchmark five algorithms to extract uncertainty estimates from deep Q-Networks. Four of them can be casted within the variational inference framework:
% that will be presented below. 
%\cite{hinton1993keeping, barber1998ensemble, graves2011practical} can be applied. Using these techniques, a simple approximate learning algorithm called Bayes By Backpropagation \cite{blundell2015weight} is obtained.

\subsubsection{Variational inference}
The intractable posterior $P(w|\mathcal{D})$ is approximated with a variational distribution $q(w|\theta)$. The parameters are learnt by minimizing the Kullback-Leibler ($\mathcal{KL}$) divergence between the variational approximation $q(w|\theta)$ and the true posterior over the weights $P(w|\mathcal{D})$. The resulting cost function is termed as the variational free energy \cite{hinton1993keeping}:
\begin{equation}
\label{eqn:freeenergy}
\mathcal{F} = \mathcal{KL}[q(w|\theta)||P(w)] -  \mathop{\mathbb{E}}_{q(w|\theta)} [\ln P(\mathcal{D}|w)].
\end{equation}
We test here four algorithms that can be casted in the variational approximate framework, namely Bayes-by-Backprop \cite{blundell2015weight,lipton2016efficient}, dropout \cite{gal2016dropout}, concrete dropout \cite{gal2017concrete} and $\alpha$-divergence \cite{li2017dropout,hernandez2016black}. 
%For the detailed explanations of the algorithms we refer to provided references.

\textbf{Deep BBQ-Learning}. We implement the Bayes By Backprop method with DQN. To propagate the error through a layer that samples from $q(w|\theta)$, the reparameterization trick is used \cite{kingma2013auto}. We choose $q(w|\theta)$ to be a Gaussian with diagonal covariance with a variational parameter set $\theta$. Given the mean  $\mu_i$ and covariance $\sigma_i$ of $q$ for each weight, a sample from q is obtained by first sampling $\epsilon_i \sim \mathcal{N}(0,\sigma_{\epsilon})$, then computing $w_i = \mu_i + \sigma_i \circ \epsilon_i$, where $\circ$ is point-wise multiplication. To ensure all $\sigma_i$ are strictly positive, the softplus function $\sigma_i = \log (1+ \exp (\rho_i ))$ is used where $\rho$ is a free parameter \cite{lipton2016efficient}. The variational parameters are then $\theta = \{{\mu_i, \rho_i}\}_{i=1}^{\mathcal{D}}$ for $D$-dimensional weight vector $w$.
The resulting gradient estimator of the variational objective is unbiased and has a lower variance. The exact cost in Eq. \ref{eqn:freeenergy} can then be approximated as:
\begin{equation}
\mathcal{F} (\mathcal{D}, \theta) \approx \sum_{i=1}^n \log q(w^{(i)}|\theta) - \log P(w^{(i)}) - \log p(\mathcal{D}|w^{(i)})
\end{equation}
where $w^{(i)}$ is the $i$th Monte Carlo sample drawn from the variational posterior $q(w^{(i)}|\theta)$.
%With minibathes drawn from the buffer, the minibatch cost for minibatch $i$ = 1, 2, ..., M is:
%\begin{equation}\label{eq:kl}
%\mathcal{F}_i (\mathcal{D}_i, \theta) = \xi_i \mathcal{KL}[q(w|\theta)||P(w)] -  \mathop{\mathbb{E}}_{q(w|\theta)} [\ln P(\mathcal{D}_i|w)]
%\end{equation}
%where $\xi_i$ is the weight for the KL cost. 
For the likelihood term in the objective function Eq. \ref{eqn:freeenergy}, we use the expected square loss. 
% as in \ref{eq:sqloss} for the likelihood term. Note that least-squares regression techniques can be interpreted as maximum likelihood with an underlying Gaussian error model. 

\textbf{$\alpha$-Divergences}. The approximate inference technique described in the Bayes By Backprop method corresponds to Variational Bayes (VB), which is a particular case of $\alpha$-divergence, where $\alpha \rightarrow 0$ \cite{hernandez2016black}. The $\alpha$-divergence measures the similarity between two distributions and can take the form:
\begin{equation}
D_{\alpha}[p||q] = {1 \over {\alpha(\alpha -1)}} (1 - \int p(\theta)^{\alpha} q(\theta) ^{1-\alpha} d \theta) ,
\end{equation}
where $\alpha \geq 0.$ 

Hernandez-Lobato et al. \cite{hernandez2016black} found that using $\alpha \neq 0$ performs better than the VB case, where an approximation with $\alpha \geq 1$ will cover all the modes of the true distribution, and the VB case only fits to a local mode, assuming the true posterior is multi-modal \cite{hernandez2016black}. $\alpha = 0.5$ achieves a balance between the two and has shown to perform best when applied to regression or classification tasks.

We experiment with an objective function based on the black box $\alpha$-divergence (BB- $\alpha$) energy. We use the reparametrization proposed by \cite{li2017dropout} :
\begin{equation}
\label{eq:bbalphaenergy}
\mathcal{L}_{\alpha} \approx \check{\mathcal{L} }_{\alpha} = \mathcal{KL}[q(w|\theta)||P(w)] - {1 \over \alpha} \sum_n \log \mathop{\mathbb{E}}_{q(w|\theta)} [P(\mathcal{D}|w)],
\end{equation}
where $\mathcal{L}_{\alpha}$ designates the BB- $\alpha$ energy, $\check{\mathcal{L} }_{\alpha}$ designates an approximation, and $n$ corresponds to the number of datapoints in the minibatch.

\textbf{Dropout}. Another method to obtain uncertainty estimates in deep neural networks is Bayesian inference with dropout \cite{gal2016dropout}. Dropout consists of randomly dropping units (with some probability $d$) from the neural network during training \cite{srivastava2014dropout}. 
%Dropout prevents overfitting and provides a way to train an ensemble of exponentially many different neural network architectures efficiently.
% As explained in \cite{li2017dropout}, dropout is equivalent to multiplying the NN weight matrices $M_i$ by some random noise $\epsilon_i$ (with a new noise realization for each data point). The resulting stochastic weight matrices $W_i = \epsilon_i M_i$ can be seen as draws from the approximate posterior over the BNN weights. 

As in the previous methods, dropout can be analyzed from the variational inference perspective (Equation \ref{eqn:freeenergy}). This comes from the fact that applying a stochastic mask is equivalent to multiplying the weight matrix in a given layer by some random noise. The resulting stochastic weight matrix can be seen as draws from the approximate posterior over weights, replacing the deterministic weight matrix \cite{gal2016dropout}. 

%The prediction uncertainty is obtained by marginalizing over the approximate posterior using Monte Carlo integration \cite{li2017dropout}:
%\begin{align}
%p(y=c|x,X,Y) & = \int p(y=c|x,w)p(w|X,Y)dw \\
%& \approx  \int p(y=c|x,w)q_{\theta}(w)dw \\
%& \approx  {1 \over K} \sum_{k=1}^K p(y=c|x, \hat{w}_k)
%\end{align}
%with $\hat{w}_k \sim q_{\theta}(w)$. 
% Dropout is a simple technique as variational approximation, without the need to make major changes to the DQN implementation. 
% Dropout variational inference is implemented by adding dropout layers before every weight layer in the neural network model. No changes to the objective function are necessary. 

\textbf{Concrete Dropout}. To  obtain  well-calibrated
uncertainty estimates with above method, a grid-search over the dropout probabilities is necessary. However, we can treat a dropout as a part of optimization task obtaining an automatic method of tuning the mask. One method is to continuously relax the dropout's discrete masks and optimize the dropout probability using gradient methods \cite{gal2017concrete}. Dropout $d$ probability becomes one of the optimized parameters. The concrete distribution relaxation $z$ of the Bernoulli random variable becomes:
$$z = \mathrm{sigmoid}(\frac{1}{t} (\log d - \log(1- d)+\log u - \log(1-u))$$
with some temperature $t$ which results in values in the interval $[0, 1 ]$ and $u \sim \mathcal{U}(0,1)$.

% \subsubsection{Noisy Networks}
% Another method 
\subsubsection{Bootstrapped DQN}
%A heuristic approach to the extraction of uncertainty estimates is the bootstrapped method \cite{osband2016deep}. 
Uncertainty estimates can be obtained with random initialization of several neural networks which predict in ensemble
%produce 
uncertainty estimates for neural networks \cite{osband2016deep}. To improve efficiency, all networks share the same architecture with a different last layer (head) computing Q-values. The algorithm obtained the highest scores in its non-bootstrapped case when all networks share the same memory replay. Here we employ this ensemble variant.

\subsubsection{Computation complexity}
To obtain uncertainty estimates GPSARSA needs $O(nk^2)$ steps, where $n$ is the total number of data points during training and $k$ is the number of representative data points ($k<<n$). Training complexity for dropout, concrete dropout and bootstrapped DQNs is $O(N)$ where $N$ is the number of neural network parameters. Complexity for BBQN is tripled as it requires three sets of parameters.

\section{Evaluation and results}
Experiments are conducted using the Cambridge restaurant domain from the PyDial toolkit \cite{ultes2017pydial} with a goal-driven user simulator on the semantic level \cite{userSim}. A user simulator replicates user behavior with sufficient accuracy to optimize model parameters to an acceptable level of performance \cite{young2013pomdp}, and is more cost-effective for development and evaluation purposes.
%then direct interaction with users, where a sufficient number of real users is usually not available. 
We use an error model where confusions to the simulated user input are added. The error model outputs an N-best list of possible user responses.
%, where the goal and the agenda are updated dynamically throughout the dialogue, and updates are controlled by decision points that can be deterministic or stochastic, enabling a wide spread of realistic dialogues to be generated \cite{gasic2014gaussian}.

The input for all models is the full dialogue belief state $b$ of size $268$ and the output action space consists of $14$ possible actions. A linear kernel was used for GPSARSA. We used DQN with two hidden layers of size $130$ and $50$. The maximum dialogue length was set to $25$ turns and $\gamma$
was $0.99$. The Adam optimiser was used with an initial learning rate of 0.001 \cite{kingma2014adam}. The results are averaged over three different runs.

\subsection{Comparison with baselines} 
\begin{figure}[!h]
\centering
  \includegraphics[width=.75\linewidth]{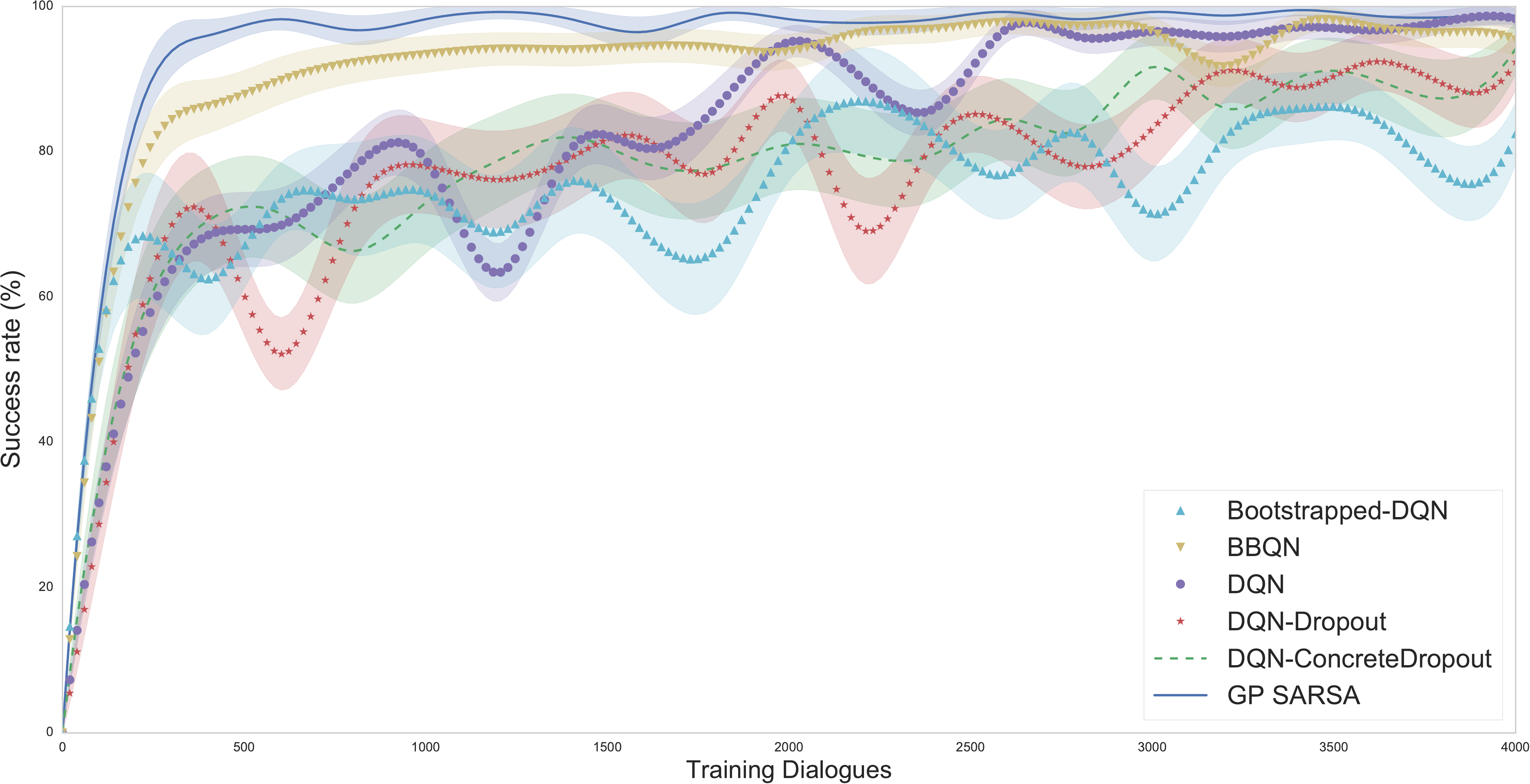}
  \caption{The success rate learning curves for all analyzed models under noise-free conditions. }
  \label{fig:mdps}
\end{figure}

Figure \ref{fig:mdps} shows learning curves for benchmarked models as a function of the training dialogues. From all analysed algorithms, only BBQN reached a performance comparable to state-of-the-art non-parametric approach in terms of the efficiency of exploration as well as the final performance. Moreover, thanks to implicit regularization due to the KL constraint, the learning becomes much more stable comparing to vanilla DQN.
%We find that GPSARSA learns the fastest and is the most stable, benefiting from the ability of Gaussian Processes to learn from a small amount of data, exploiting the correlations defined by the kernel function (Fig. \ref{fig:mdps}, left). 

Three other analyzed methods, dropout, concrete dropout and bootstrapped approach, did not help improving learning rate over the vanilla $\epsilon$-greedy algorithm neither do they stabilize exploration. 
Although with concrete dropout tuning of the dropout probability is automatic, it did not help improve efficiency.
We also optimize over number of heads with bootstrapped DQN, however, the performance did not vary substantially yielding the best results with $5$ heads.

For $\alpha$-divergences, we find all settings of $\alpha$ do not perform better than VI in general, for $k > 1$ samples. For clarity we did not show here learning curves. Taking more MC samples decreases the variance of the gradient estimates, and the averaged loss for most updates is closer to the loss obtained when taking a sample close to the mean of the variational distribution $q$. This implies more updates are necessary to move in the direction of the true posterior distribution $p$, resulting in slower convergence to an optimal policy. 
%(by minimizing the $\mathcal{KL}$ or $\alpha$ divergence), 
%trading off for reduced exploration, and slower learning. %With the BB-$\alpha$ reparametrization, more than $1$ samples are necessary for $\alpha$-divergence to have any effect, but this reduces exploration efficiency and results in worse performance.

\subsection{Noise-robustness}
\begin{figure}[!h]
\centering
 \includegraphics[width=.75\linewidth]{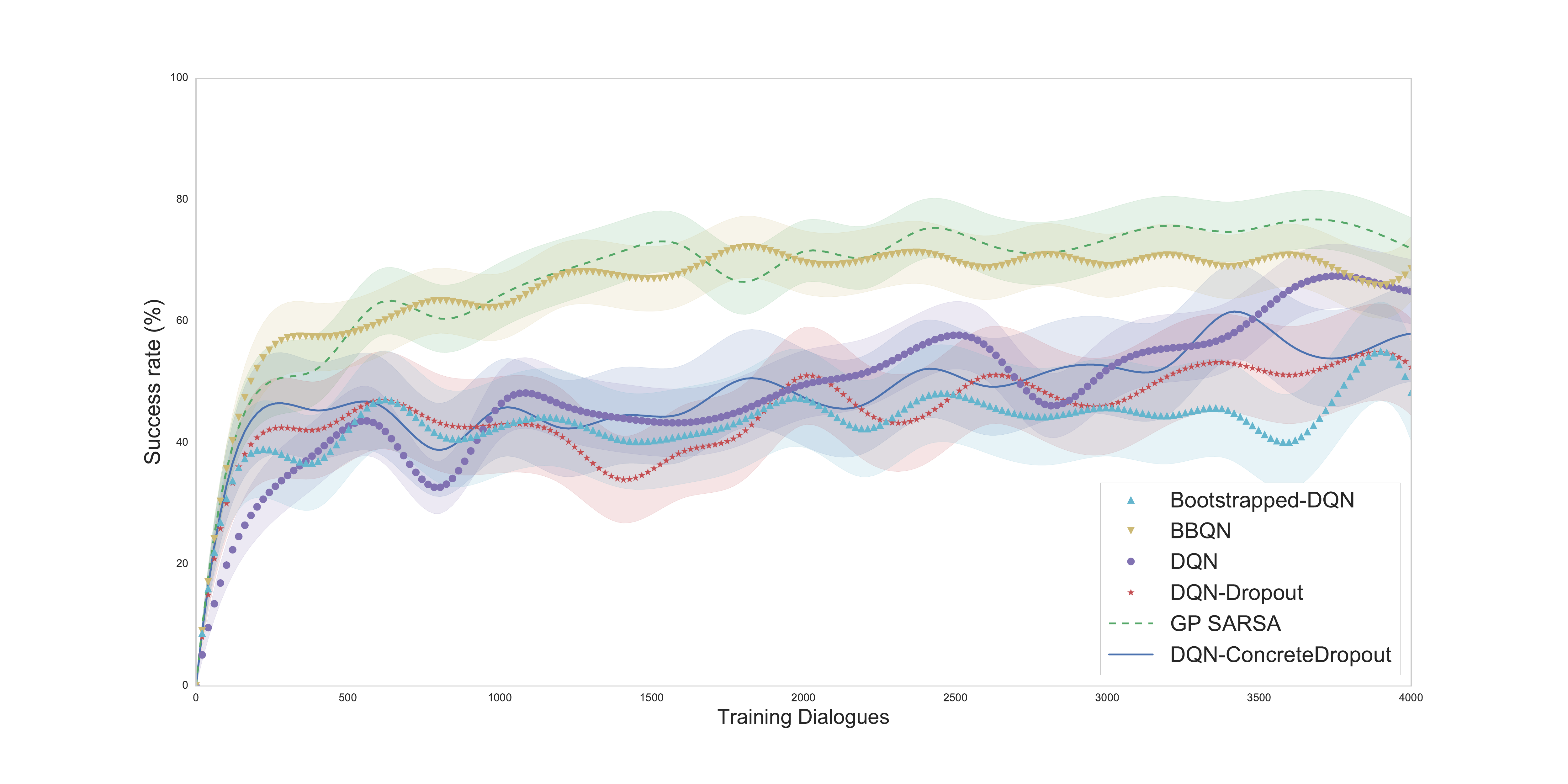}
  \caption{The success rate learning curves for all analyzed models with a  45\% confusion rate at testing, and 15\% confusion rate during training. }
  \label{fig:mdps2}
\end{figure}

We also investigated the impact of noise by training all models with the simulated user with a 15\% semantic error rate, then evaluated on 45\% semantic error rate to examine the generalisation capabilities of different algorithms. The final success rates are given in Figure \ref{fig:mdps2} as a function of the training dialogues.
  
The results show that GPSARSA performs best in terms of success rate, followed closely by BBQN. This shows that BBQN generalizes better than $\varepsilon$-greedy algorithms. BBQN has the potential for robust performance, and performs well, even at conditions different from the training conditions. All other methods fall behind substantially with only vanilla DQN being able to reach similar performance at the end of the training. 

%%%%%%%EPSRC
%\section*{Acknowledgments}
%This work is partially funded by
% thanks to dialogue group for suggestions

\newpage
\bibliography{report}
\bibliographystyle{plainnat}
\end{document}